\documentclass[11pt,letterpaper]{article}
\usepackage[T1]{fontenc}
\usepackage[utf8]{inputenc}
\usepackage{amsmath,mathtools}
\usepackage{libertine}
\usepackage[libertine]{newtxmath}
\usepackage[margin=1in]{geometry}
\usepackage{microtype,booktabs,array,tabularx,graphicx,xcolor}
\usepackage{caption,enumitem,tikz,needspace,placeins,titlesec}
\usepackage{pgfplots}
\pgfplotsset{compat=1.18}
\usetikzlibrary{arrows.meta,positioning}
\usepackage[backend=biber,style=alphabetic,sorting=nyt,maxbibnames=6,minbibnames=3,giveninits=true]{biblatex}
\usepackage[colorlinks=true,linkcolor=blue,citecolor=blue,urlcolor=blue]{hyperref}
\usepackage[nameinlink,noabbrev]{cleveref}

\definecolor{ink}{RGB}{35,52,70}
\definecolor{panel}{RGB}{245,247,249}
\newcolumntype{Y}{>{\raggedright\arraybackslash\hyphenpenalty=10000\exhyphenpenalty=10000}X}
\newcolumntype{L}[1]{>{\raggedright\arraybackslash\hyphenpenalty=10000\exhyphenpenalty=10000}p{#1}}
\titleformat{\section}{\Large\bfseries}{\thesection}{0.7em}{}
\titleformat{\subsection}{\large\bfseries}{\thesubsection}{0.7em}{}
\titlespacing*{\section}{0pt}{13pt plus 2pt minus 1pt}{6pt}
\titlespacing*{\subsection}{0pt}{10pt plus 2pt minus 1pt}{4pt}
\AtBeginBibliography{\small\interlinepenalty=10000\urlstyle{same}}
\hypersetup{pdftitle={WenSight-Decision-1 for Direct Decisions under E-commerce Rules},pdfauthor={Mingwen Liu; Jun Sun; Sichen Wang}}

\newcommand{\authorblock}[3]{%
  \begin{minipage}[t]{0.32\linewidth}
    \centering
    {\fontsize{12}{15}\selectfont #1\par}\vspace{3pt}%
    {\footnotesize\href{mailto:#2}{\color{black}\texttt{#2}}\par}\vspace{2pt}%
    {\footnotesize #3\par}%
  \end{minipage}}

\title{\fontsize{15.5}{20}\selectfont WenSight-Decision-1 for Direct Decisions under E-commerce Rules}
\author{%
  \authorblock{Mingwen Liu}{maxwell@alphafuture.cn}{WenSight AI Lab}\hfill
  \authorblock{Jun Sun}{290060568@qq.com}{TooChuang}\hfill
  \authorblock{Sichen Wang}{sichenwang@cuhk.edu.cn}{WenSight AI Lab}}
\date{}

\makeatletter
\renewcommand{\maketitle}{%
  \par\vspace*{6pt}\begingroup\centering
  {\@title\par}\vspace{18pt}%
  \begin{minipage}{0.86\textwidth}\centering\@author\end{minipage}\par%
  \endgroup\vspace{18pt}}
\makeatother
\renewenvironment{abstract}{%
  \par\begingroup\small
  \centerline{\bfseries Abstract}\vspace{5pt}%
  \begin{list}{}{\setlength{\leftmargin}{24pt}\setlength{\rightmargin}{24pt}%
    \setlength{\topsep}{0pt}\setlength{\partopsep}{0pt}\setlength{\parsep}{0pt}}%
  \item\relax
}{\end{list}\endgroup\vspace{4pt}}

\begin{document}
\maketitle
\begin{abstract}
E-commerce decisions require applying natural-language rules to the available case evidence. We introduce WenSight-Decision-1, a shared model that scores supplied options directly from rules and case information for six operational tasks, with examples spanning six languages. We fine-tune Qwen3.5-9B with reference options as supervision, using the same option scores for training and inference. On 1,440 synthetic operational cases reserved for final evaluation, accuracy rises from 61.60\% for Qwen3.5-9B to 92.92\% for WenSight-Decision-1 under identical inputs and scoring. Joint fine-tuning improves all six tasks, bringing heterogeneous operational judgments under a common decision interface.
\end{abstract}

\section{Introduction}
\label{sec:introduction}

Approving a refund, routing a customer request, and reviewing a product claim require different judgments, each governed by operational rules and case evidence. ShARC studies how natural-language rules are interpreted against case information, including when clarification is needed~\cite{saeidi2018sharc}; RuleArena documents continuing difficulties in applying complex rules from practical domains~\cite{zhou2025rulearena}. Such judgments also form part of the retail interactions studied by $\tau$-bench and Mix-Ecom~\cite{yao2025taubench,zhou2025mixecom}. We study six e-commerce tasks arising from the operational needs of our company or its customers. The central problem is to make these judgments with one model while allowing the rules and available answers to vary from case to case.

WenSight-Decision-1 unifies these tasks in a shared model based on \mbox{Qwen3.5-9B}~\cite{qwen35}. Each input supplies a question, operational rules, case information, and descriptions of the available options. The model scores all options in one forward pass and returns the highest-scoring answer, giving a System~1-style decision path. Supervised fine-tuning uses the same option scores to favor the reference answer. Task-specific criteria and option meanings thus reside in the input, while scoring and supervision follow the same formulation across tasks. WenSight-Decision-1 is the first multi-task System~1-style model developed specifically for decisions governed by e-commerce operational rules.

On 1,440 synthetic operational cases reserved for final evaluation, accuracy rises from 61.60\% for the starting Qwen3.5-9B checkpoint to 92.92\% for WenSight-Decision-1 under identical inputs and scoring. The overall gain is 31.32 percentage points, and every task improves. Paired development tests examine stability under equivalent wording and correctness when the expected answer changes. The same formulation supports supervised adaptation to other tasks expressed through rules, case information, and a finite set of options.

\Needspace{6\baselineskip}
\subsection{Related Work}
\label{sec:related-work}

Rule interpretation and policy-conditioned judgments appear in several settings. ShARC emphasizes conversational clarification~\cite{saeidi2018sharc}, RuleTaker studies what follows from textual facts and rules~\cite{clark2020ruletaker}, and RuleArena addresses complex regulations and computations~\cite{zhou2025rulearena}. For safety assessment, SingGuard accepts an active policy and supports direct judgments or extended reasoning~\cite{singguard2026}; ToolGuard compiles company policy into code that checks tool use before execution~\cite{zwerdling2025toolguard}. WenSight-Decision-1 learns to select among supplied operational options, including information requests or human review when the rules require them.

A common choice format allows one model to serve multiple tasks. UniMC uses masked option representations and option-prediction supervision~\cite{yang2022unimc}; GLiClass jointly encodes text and dynamic labels for scoring in one forward pass~\cite{stepanov2025gliclass}. Dyad uses a separate action encoder and learns through environment interaction~\cite{zhan2026dyad}. WenSight-Decision-1 uses the next-token scores of letters assigned to options and supervised fine-tuning on reference choices. A similar readout appears in the official Simple Jev technical documentation~\cite{featherlesssimplejev}.

EcomGPT and eCeLLM study multi-task instruction tuning for e-commerce~\cite{li2024ecomgpt,peng2024ecellm}, while $\tau$-bench and Mix-Ecom evaluate interactions involving conversations, tools, and domain rules~\cite{yao2025taubench,zhou2025mixecom}. WenSight-Decision-1 addresses the rule-conditioned choices within these settings through a shared model for six operational tasks.

\section{Policy-Conditioned Decision Tasks}
\label{sec:decision-tasks}
\label{sec:running-example}

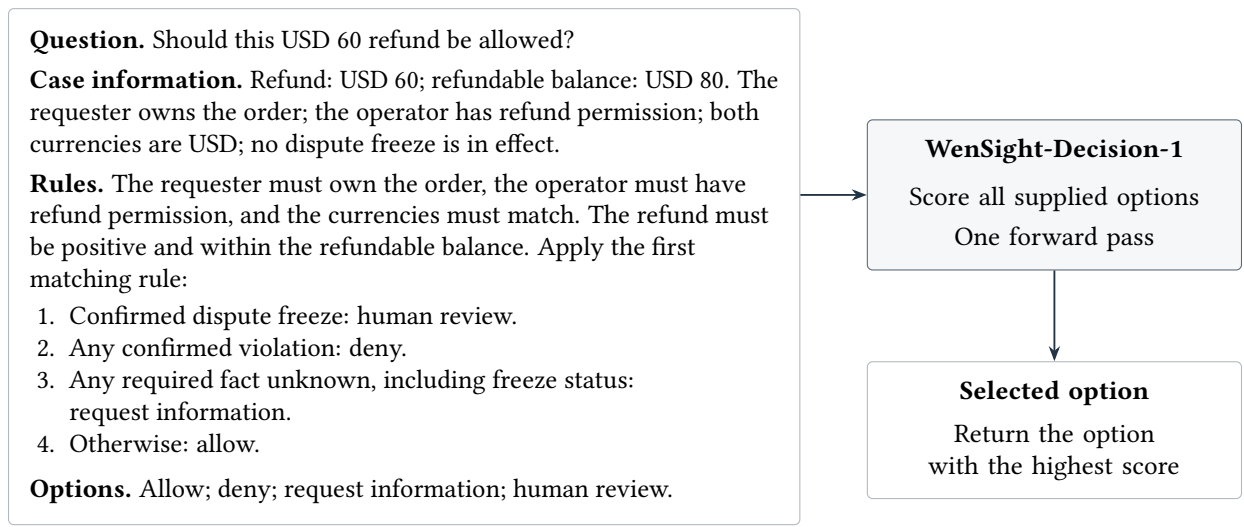
\begin{figure}[htbp]
\centering
\begin{tikzpicture}[
  inputbox/.style={draw=ink!35,rounded corners=2pt,inner sep=8pt},
  modelbox/.style={draw=ink!75,fill=panel,rounded corners=2pt,
    text width=0.27\textwidth,align=center,inner sep=7pt,font=\small},
  outputbox/.style={draw=ink!35,rounded corners=2pt,
    text width=0.27\textwidth,align=center,inner sep=7pt,font=\small},
  flow/.style={-{Stealth[length=2mm]},draw=ink,line width=0.7pt}
]
\node[inputbox,anchor=north west] (input) {%
\begin{minipage}{0.60\textwidth}
\small
\raggedright\hyphenpenalty=10000\exhyphenpenalty=10000
\textbf{Question.} Should this USD~60 refund be allowed?\par\smallskip
\textbf{Case information.} Refund: USD~60; refundable balance: USD~80. The requester owns the order; the operator has refund permission; both currencies are USD; no dispute freeze is in effect.\par\smallskip
\textbf{Rules.} The requester must own the order, the operator must have refund permission, and the currencies must match. The refund must be positive and within the refundable balance. Apply the first matching rule:
\begin{enumerate}[label=\arabic*.,leftmargin=1.5em,topsep=2pt,itemsep=0pt,parsep=0pt]
\item Confirmed dispute freeze: human review.
\item Any confirmed violation: deny.
\item Any required fact unknown, including freeze status: \mbox{request information}.
\item Otherwise: allow.
\end{enumerate}
\smallskip
\textbf{Options.} Allow; deny; request information; human review.
\end{minipage}};
\node[modelbox,anchor=west] (model) at ([xshift=25pt,yshift=27pt]input.east)
  {\textbf{WenSight-Decision-1}\\[6pt]Score all supplied options\\[3pt]One forward pass};
\node[outputbox,below=12mm of model] (choice)
  {\textbf{Selected option}\\[4pt]Return the option with the highest score};
\draw[flow] (input.east |- model.west) -- (model.west);
\draw[flow] (model) -- (choice);
\node[anchor=north west,inner sep=0pt,font=\small,align=left]
  at ([yshift=-8pt]input.south west)
  {Expected answer under these rules: \textbf{allow}.};
\end{tikzpicture}
\caption{A refund-review example and the direct option-scoring path.}
\label{fig:decision-example}
\end{figure}

The refund review in \Cref{fig:decision-example} shows how rules and evidence determine a decision. A USD~60 refund against a refundable balance of USD~80 is allowed when the ownership, permission, currency, and dispute checks are satisfied. Changing only the permission status gives two other outcomes: unknown permission calls for information, while a confirmed lack of permission requires denial. A confirmed dispute freeze calls for human review.

\label{sec:task-formulation}
We call choosing one supplied answer according to operational rules and available case information \emph{policy-conditioned decision making}. A \emph{policy} is the text specifying the decision criteria. A \emph{rule} is an individual requirement within it. An input consists of a question $q$, policy text $r$, case information $s$, and a nonempty finite list of options $O=(o_1,\ldots,o_K)$, where $K$ is the number of options. The task permits incomplete case information in $s$.

Each option $o_j$ has an identifier and a textual description, with $j\in\{1,\ldots,K\}$ denoting its position in the presented order. For example, \texttt{allow} identifies the option \emph{Approve the proposed refund}. Each input specifies its own options and their meanings. For a labeled case, the \emph{reference label} $y\in\{1,\ldots,K\}$ is the position of the annotated option $o_y$; the label and any accompanying annotation are excluded from the model input. The model selects a position $\hat y\in\{1,\ldots,K\}$ and returns $o_{\hat y}$. A prediction is correct when $\hat y=y$, and reordering options updates $y$ to the annotated option's new position.

We use \emph{decision} for the selected answer and \emph{action} for a concrete operation such as issuing a refund. The six tasks share this input structure but differ in their judgments and option meanings. Their examples cover English, Chinese, German, Spanish, French, and Japanese.

\emph{Support routing} assigns requests to queues according to responsibilities, priority, and ownership. \emph{Product matching} checks identity or attribute requirements, with options such as a match, a non-match, and insufficient information. \emph{Listing review} judges whether a product claim is supported, unsupported, or requires review.

\emph{Priority classification} assigns an urgency level from P0 to P3, with higher numbers indicating greater urgency in our data. \emph{Next-step selection} chooses whether to answer, look up information, request clarification, or refer for human handling. \emph{Action review} checks permissions and eligibility for a proposed operation, as in the refund example.

\FloatBarrier
\section{Model and Fine-Tuning}
\label{sec:model}

\subsection{Scoring and Supervision}
\label{sec:option-scoring}
\label{sec:supervised-fine-tuning}

We format $q$, $r$, $s$, and $O$ as a text prompt, assigning each option a distinct single-letter label alongside its description. Let $x$ denote the tokenized input, $\theta$ the model's full effective parameters, and $c_j$ the token ID of the letter assigned to $o_j$. We encode options with the single-token letters A--Z, supporting $1\leq K\leq26$ options and inputs of at most 2,048 tokens.

Let $z_\theta(x)$ be the vector of unnormalized next-token scores, or logits, at the final input position, and define the option logit $z_j=z_\theta(x)_{c_j}$. \emph{Direct option scoring} selects an option from these logits without generating answer text or a reasoning trace. Normalizing over the current options gives the option probability $p_\theta(j\mid x)$, while the largest logit determines the predicted position:
\[
p_\theta(j\mid x)=\frac{\exp(z_j)}{\sum_{k=1}^{K}\exp(z_k)},
\qquad
\hat y=\underset{1\leq j\leq K}{\operatorname{argmax}}\;z_j.
\]
All option logits come from one forward pass. Ties favor the first option in the presented order, and the selected position maps back to $o_{\hat y}$. With explicit thinking disabled in the prompt template, this single-pass selection gives a \emph{System~1-style} decision path.

The same logits support supervised fine-tuning. For a labeled example $(x,y)$, the single-example loss $\mathcal{L}(\theta;x,y)$ is cross-entropy over the options:
\[
\mathcal{L}(\theta;x,y)=-\log p_\theta(y\mid x).
\]
During training, we randomly permute the options and update the reference position and letter mapping together. Task-specific rules and option meanings enter through the input, so all tasks share one readout and loss. The same formulation applies to internal request routing, for example, with routing rules, department descriptions as options, and labeled routing decisions for fine-tuning.

\subsection{Model Adaptation}
\label{sec:training-data}

We fine-tune the released, post-trained Qwen3.5-9B checkpoint~\cite{qwen35} on text inputs using low-rank adaptation (LoRA)~\cite{hu2021lora}. We train approximately 43.28 million adapter parameters in the text modules, or 0.46\% of the loaded model's parameters, while keeping the original weights frozen.

The operational data have three synthetic sources. \emph{Initial synthetic examples} are AI-authored cases with reference answers determined from the supplied rules and case information; they are used for training, development, and final evaluation. \emph{Separately authored synthetic examples} are used only for development and final evaluation, with the development portion contributing to checkpoint selection. \emph{Harder synthetic examples} were designed for more involved rule and evidence judgments, including permission preconditions and claim support. Their preparation included targeted AI semantic review of rule and label consistency. An auxiliary \emph{ESCI subset} comes from Amazon's Shopping Queries Dataset~\cite{reddy2022esci}. We retain its published human labels (Exact, Substitute, Complement, and Irrelevant) for search relevance classification in training, development, and final evaluation.

The first fine-tuning stage uses initial synthetic examples and the ESCI subset. The second continues from the first-stage adapter, adding harder synthetic examples while replaying earlier training data. Harder examples enter only second-stage training and development. The stages use approximately 4,000 and 6,000 training records, respectively; the second-stage development set contains 1,311 examples. The final checkpoint is selected by development-set performance. Final evaluation counts by source appear in \Cref{tab:final-performance}.

\section{Evaluation}
\label{sec:evaluation}

We compare WenSight-Decision-1 with its starting Qwen3.5-9B checkpoint, the \emph{reference model}, using identical inputs and direct option scoring. This comparison measures the gain from fine-tuning. We fix the checkpoint and scoring configuration before final evaluation.

The final evaluation contains 1,440 synthetic operational cases and 200 auxiliary ESCI cases, held out from training and model selection. \emph{Accuracy} is the fraction of cases whose selected semantic option matches the reference answer, aggregated by case; $N$ denotes the number of cases in a reported group. Behavioral tests use development cases.

\subsection{Decision Performance}
\label{sec:decision-performance}

Across the 1,440 operational cases, accuracy rises from 61.60\% to 92.92\%, a gain of 31.32 percentage points. \Cref{tab:final-performance} separates the results by task and source.

\begin{table}[htbp]
\centering
\small
\caption{Final evaluation by task (A) and source (B). Accuracy pairs are Qwen3.5-9B $\rightarrow$ WenSight-Decision-1 (\%). Panel B includes 200 auxiliary ESCI cases in addition to the 1,440 operational cases.}
\label{tab:final-performance}
\setlength{\tabcolsep}{3pt}
\begin{minipage}[t]{0.50\textwidth}
\textbf{(A) Operational tasks}\par\vspace{4pt}
\begin{tabularx}{\linewidth}{@{}Yr r@{\,$\rightarrow$\,}r@{}}
\toprule
\textbf{Task} & $N$ & \multicolumn{2}{c@{}}{\textbf{Accuracy (\%)}} \\
\midrule
Support routing & 235 & 68.94 & \textbf{87.23} \\
Product matching & 241 & 84.23 & \textbf{96.68} \\
Listing review & 241 & 57.68 & \textbf{95.02} \\
Priority classification & 241 & 49.79 & \textbf{96.68} \\
Next-step selection & 241 & 54.77 & \textbf{94.19} \\
Action review & 241 & 54.36 & \textbf{87.55} \\
\midrule
\textbf{Operational total} & \textbf{1,440} & \textbf{61.60} & \textbf{92.92} \\
\bottomrule
\end{tabularx}
\end{minipage}\hfill
\begin{minipage}[t]{0.47\textwidth}
\textbf{(B) Data sources}\par\vspace{4pt}
\begin{tabularx}{\linewidth}{@{}Yr r@{\,$\rightarrow$\,}r@{}}
\toprule
\textbf{Source} & $N$ & \multicolumn{2}{c@{}}{\textbf{Accuracy (\%)}} \\
\midrule
Initial synthetic & 1,200 & 62.17 & \textbf{96.08} \\
Separately authored & 240 & 58.75 & \textbf{77.08} \\
ESCI (auxiliary) & 200 & 43.50 & \textbf{51.50} \\
\midrule
\textbf{All cases} & \textbf{1,640} & \textbf{59.39} & \textbf{87.87} \\
\bottomrule
\end{tabularx}
\end{minipage}
\end{table}

Fine-tuning improves all six operational tasks, with the largest gain in Priority classification. The gains extend to both synthetic sources, with accuracy reaching 96.08\% on the initial examples and 77.08\% on the separately authored cases. On the auxiliary ESCI subset, accuracy rises from 43.50\% to 51.50\%.

\FloatBarrier

\Needspace{6\baselineskip}
\subsection{Behavior and Latency}
\label{sec:behavioral-evaluation}

Paired tests examine how decisions respond to controlled changes in the input. Following CheckList~\cite{ribeiro2020checklist}, \emph{invariance tests} preserve the expected semantic answer. \emph{Label-changing edits} make a different answer correct under the rules. Each pair contains an original input and a transformed version. We compare predictions by option meaning: the \emph{prediction flip rate} is the fraction of pairs with different semantic predictions. A pair has \emph{both predictions correct} when each prediction matches its own reference answer; we report the fraction of such pairs.

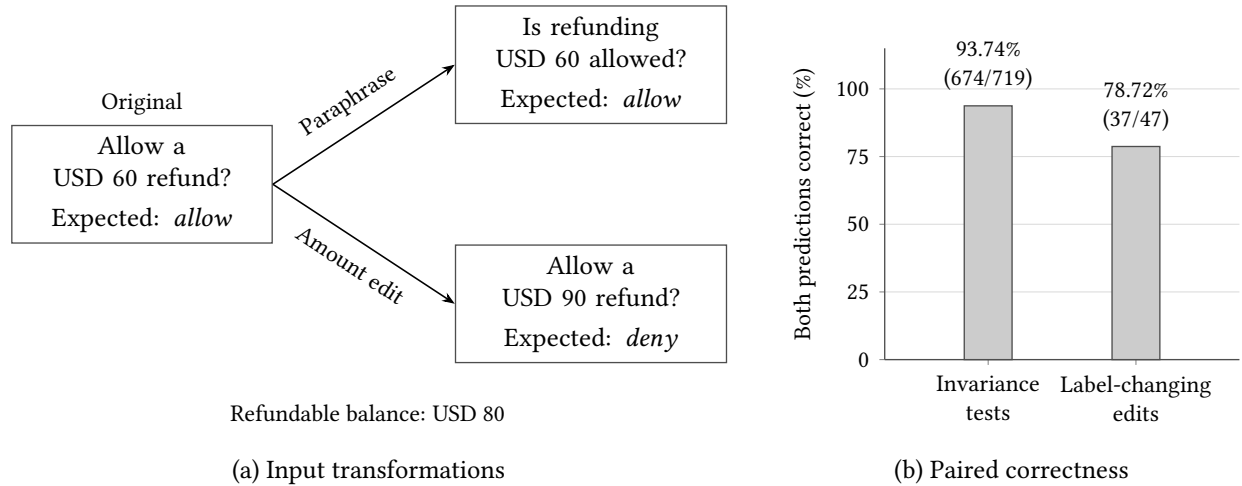
\begin{figure}[htbp]
\centering
\begin{minipage}[b]{0.59\textwidth}
\centering
\begin{tikzpicture}[
  case/.style={draw=black!70,line width=0.5pt,rectangle,
    text width=90pt,minimum height=44pt,inner sep=4pt,
    align=center,font=\small},
  edit/.style={-{Stealth[length=1.7mm]},line width=0.6pt},
  every label/.style={font=\footnotesize}
]
\node[case,label=above:Original] (original) at (0,0)
  {Allow a USD~60 refund?\\[3pt]Expected: \emph{allow}};
\node[case,text width=94pt] (paraphrase) at (169pt,45pt)
  {Is refunding USD~60 allowed?\\[3pt]Expected: \emph{allow}};
\node[case,text width=94pt] (amount) at (169pt,-45pt)
  {Allow a USD~90 refund?\\[3pt]Expected: \emph{deny}};
\draw[edit] (original.east) -- node[sloped,above=2pt,font=\footnotesize] {Paraphrase} (paraphrase.west);
\draw[edit] (original.east) -- node[sloped,below=2pt,font=\footnotesize] {Amount edit} (amount.west);
\node[font=\footnotesize,anchor=north] at (85pt,-79pt)
  {Refundable balance: USD~80};
\end{tikzpicture}\par\vspace{4pt}
{\small (a) Input transformations}
\end{minipage}\hfill
\begin{minipage}[b]{0.38\textwidth}
\centering
\pgfmathsetmacro{\invarianceAccuracy}{674/719*100}
\pgfmathsetmacro{\editAccuracy}{37/47*100}
\begin{tikzpicture}
\begin{axis}[
  width=\linewidth,height=57mm,
  ybar,bar width=18pt,
  ymin=0,ymax=115,
  ytick={0,25,50,75,100},
  xtick={1,2},
  xticklabels={{Invariance\\tests},{Label-changing\\edits}},
  xticklabel style={align=center,font=\footnotesize},
  yticklabel style={font=\footnotesize},
  ylabel={Both predictions correct (\%)},
  ylabel style={font=\footnotesize},
  axis lines*=left,
  xmin=0.3,xmax=2.7,enlarge x limits=false,
  axis line style={black!70,line width=0.5pt},
  tick align=outside,major tick length=2pt,
  ymajorgrids=true,grid style={black!15,line width=0.25pt},
  point meta=explicit symbolic,
  nodes near coords,
  every node near coord/.append style={font=\footnotesize,align=center,anchor=south,yshift=2pt},
  clip=false
]
\addplot[fill=black!20,draw=black!70,line width=0.5pt] coordinates {
  (1,\invarianceAccuracy) [93.74\%\\(674/719)]
  (2,\editAccuracy) [78.72\%\\(37/47)]
};
\end{axis}
\end{tikzpicture}\par\vspace{4pt}
{\small (b) Paired correctness}
\end{minipage}
\caption{Paired behavioral tests. (a) Two edits of the refund case in \Cref{fig:decision-example}; all other facts remain fixed. Node labels give the expected answers. (b) Percentages and counts of pairs with both predictions correct, measured on separate development sets under the checkpoint-selection scoring configuration. Semantic predictions change in 7 of 719 paraphrase pairs (0.97\%).}
\label{fig:behavioral-evaluation}
\end{figure}

\Cref{fig:behavioral-evaluation} illustrates both transformations using the refund example and reports the selected model's paired development results.

\FloatBarrier

Under the final scoring configuration, WenSight-Decision-1 answers 113 of 120 original development cases correctly. Varying option order, option identifiers, and spacing in option descriptions produces 365 answer-preserving pairs with changed model inputs; one pair has a semantic prediction flip.

We measure \emph{round-trip latency}, from request to response, with a resident service on a DGX Spark with one NVIDIA GB10 and a 2,048-token input limit. After warm-up, we send 614 requests from the same 120 cases and their variants serially over fresh local loopback connections. Median (p50) latency is 356.48~ms, and the 95th percentile (p95) is 514.38~ms.

\section{Conclusion}
\label{sec:conclusion}

WenSight-Decision-1 brings six rule-conditioned e-commerce tasks into one model that selects among supplied options in a single forward pass. Rules and option meanings enter through the input, while the scoring rule and supervision objective are shared across tasks. Joint fine-tuning raises operational accuracy from 61.60\% to 92.92\% on the reserved evaluation, with improvements across all six tasks. The shared input format, readout, and training objective support supervised adaptation to other decisions expressed through textual rules, case information, and finite option sets.

\clearpage
\printbibliography
\end{document}